\documentclass[sigconf,9pt]{acmart}
\hypersetup{draft}
\usepackage{booktabs} % For formal tables
\usepackage{hyperref}
\usepackage{lipsum}
\usepackage{graphicx}
\usepackage{siunitx, amsmath}
\usepackage{subfigure}
\usepackage{pifont}% http://ctan.org/pkg/pifont
\usepackage[normalem]{ulem}
\usepackage{multirow}
\usepackage{multicol}
\usepackage{comment}
\usepackage{color}
\usepackage{etoolbox}
\newbool{inccomment}
\setbool{inccomment}{true}
\newcommand{\hl}[1]{\ifbool{inccomment}{{\color{magenta}#1}}{}}
\newcommand{\fc}[1]{\ifbool{inccomment}{{\color{blue}#1}}{}}
\newcommand{\BiState}[2]{%
  \State{\makebox[2.4cm]{#1\hfill}#2}%
  }

\usepackage{bbm}

\usepackage{tabularx}
\usepackage{graphicx}
\usepackage{supertabular}
\usepackage{textcomp}

\usepackage{algorithm}
\usepackage{algpseudocode}
\algtext*{EndWhile}
\algtext*{EndIf}
\algtext*{EndFor}
\algtext*{EndProcedure}
\usepackage[super]{nth}
\newcommand{\thistheoremname}{}
\newtheorem*{genericthm*}{\thistheoremname}
\newenvironment{namedthm*}[1]
  {\renewcommand{\thistheoremname}{#1}%
   \begin{genericthm*}}
  {\end{genericthm*}}
 
\usepackage{mathtools}

\usepackage{url}

\graphicspath{{./_fig/}}

\usepackage{amsthm}

\allowdisplaybreaks
\makeatletter
\def\BState{\State\hskip-\ALG@thistlm}
\makeatother
\copyrightyear{2021} 
\acmYear{2021} 
\setcopyright{acmcopyright}\acmConference[ASPDAC '21]{26th Asia and South Pacific Design Automation Conference}{January 18--21, 2021}{Tokyo, Japan}
\acmBooktitle{26th Asia and South Pacific Design Automation Conference (ASPDAC '21), January 18--21, 2021, Tokyo, Japan}
\acmPrice{15.00}
\acmDOI{10.1145/3394885.3431562}
\acmISBN{978-1-4503-7999-1/21/01}

\begin{document}

%\title{WireGraph: A Customized Graph Neural Network Method for Pre-Placement Net Length Estimation}

%\title{\vspace{-2mm} Net$^\text{2}$: A Graph Attention \underline{Net}work Method \\ Customized for Pre-Placement \underline{Net} Length Estimation \vspace{-3mm}}

\title{Net$^\text{2}$: A Graph Attention \underline{Net}work Method \\ Customized for Pre-Placement \underline{Net} Length Estimation \vspace{-3mm}}

\author{Zhiyao Xie$^\dagger$, Rongjian Liang$^\mathsection$, Xiaoqing Xu$^\ddagger$, Jiang Hu$^\mathsection$, Yixiao Duan$^\dagger$, Yiran Chen$^\dagger$}

\affiliation{
\institution{$^\dagger$ Duke University, Durham, NC, USA}
\vspace{0.5mm}
\institution{$^\ddagger$ ARM Inc., Austin, TX, USA}
\vspace{0.5mm}
\institution{$^\mathsection$ Texas A\&M University, College Station, TX, USA}
\vspace{1mm}
\institution{zhiyao.xie@duke.edu  \; \; \;  yiran.chen@duke.edu}
}

%\title{NetNet: A Customized Graph Neural \underline{Net}work Method for Pre-Placement \underline{Net} Length Estimation}

%\author{Anonymous authors\vspace{5mm}}

%=======================================%
%=======================================%

%=======================================%
%               Sections                %
%=======================================%
%=======================================%
%           Abstract          %
%=======================================%
\begin{abstract}
Net length is a key proxy metric for optimizing timing and power across various stages of a standard digital design flow.
However, the bulk of net length information is not available until cell placement, and hence it is a significant challenge to explicitly consider net length optimization in design stages prior to placement, such as logic synthesis.
This work addresses this challenge by proposing a graph attention network method with customization, called Net$^\text{\textbf{2}}$, to estimate individual net length before cell placement. Its accuracy-oriented version Net$^\text{\textbf{2a}}$ achieves about 15\% better accuracy than several previous works in identifying both long nets and long critical paths. Its fast version Net$^\text{\textbf{2f}}$ is more than 1000$\times$ faster than placement while
still outperforms previous works and other neural network techniques in terms of various accuracy metrics.

%by capturing rich global information with a flexible graph model.For different application scenarios, we provide a faster version Net$^\text{\textbf{2f}}$ and a more accurate version . Net$^\text{\textbf{2f}}$ is more than 1000 $\times$ faster than placement while Net$^\text{\textbf{2a}}$ 
%Experimental results show that Net$^\text{\textbf{2}}$   achieves 10-15\% better accuracy than previous works in identifying both long nets and long critical paths.

\iffalse
Wirelength is a key factor that shapes chip timing and power, and needs to be considered in early design steps although it is not decided until layout. A customized graph neural network (GNN) method named Net$^\text{\textbf{2}}$ is developed for estimating individual net length prior to cell placement. It achieves high accuracy by capturing rich global information with a flexible graph model. Experimental results show that Net$^\text{\textbf{2}}$ remarkably outperforms previous works as well as plug-in use of existing GNN techniques. It achieves 10-15\% better accuracy than previous works in identifying both long nets and long critical paths.
\fi

\end{abstract}

\maketitle

\section{Introduction}

For state-of-the-art semiconductor manufacturing technology nodes, interconnect is a dominating factor for integrated circuit (IC) performance and power, e.g., it can contribute to over 1/3 of clock period~\cite{hyun2019accurate} and about 1/2 of total chip dynamic power~\cite{MagenSLIP04}. Interconnect characteristics are affected by almost every step in a design flow, but not explicitly quantified and optimized until the layout stage. 
Therefore, previous academic studies attempted to address the interconnect effect in design steps prior to layout, e.g., layout-aware synthesis~\cite{PedramDAC91,PedramICCAD91}. 
Moreover, a recent industrial trend among commercial implementation tools~\cite{FusionCompiler,ispatial} takes an ambitious goal to explicitly address the interaction between logic synthesis and layout.
To achieve such a goal, an essential element is to enable fast yet accurate pre-layout net length prediction, which has received significant research attention in the past\cite{bodapati2001prelayout,fathi2009pre,liu2012neural,hu2003wire,kahng2005intrinsic,liu2004pre,hyun2019accurate}. Some works \cite{bodapati2001prelayout,fathi2009pre} pre-define numerous features describing each net, then a polynomial model is built by fitting these features.
The work of \cite{liu2012neural} estimates wirelength by artificial neural networks (ANN), but it is limited to the total wirelength on an FPGA only, which is easier to estimate than individual net length. The mutual contraction (MC) \cite{hu2003wire} estimates net length by checking the number of cells in every neighboring net. The intrinsic shortest path length (ISPL) \cite{kahng2005intrinsic} is an interesting heuristic, which finds the shortest path between cells in the net to be estimated, apart from the net itself. The idea in \cite{liu2004pre} is similar to \cite{kahng2005intrinsic} in measuring the graph distance between cells in the netlist. The recent work \cite{hyun2019accurate} on wirelength prediction can only estimate the wirelength of an entire path instead of individual nets, and it relies on the results from virtual placement and routing.

Although net length prediction has been extensively studied previously, we notice a major limitation in most works.
That is, they only focus on the local topology around each individual net with an over-simplified model. In other words, when estimating each net, usually their features only include information from nets one or two-hop away. The big picture, which is the net's position in the whole netlist, is largely absent. However, a placer normally optimizes a cost function defined on the whole netlist. It is not likely to achieve high accuracy without accessing any global information. Some previous models indeed attempt to embrace global information like the number of 2-pin nets in an entire circuit~\cite{bodapati2001prelayout,fathi2009pre}, or a few shortest paths~\cite{kahng2005intrinsic}, but such information is either too sketchy~\cite{bodapati2001prelayout,fathi2009pre} or still limited to a region of several hops~\cite{kahng2005intrinsic}. 
Since the global or long-range impact on individual nets is much more complex than local circuit topologies, it can hardly be captured by simple models or models with only human-defined parameters that cannot learn from data. 
To solve this, we propose a new approach, called Net$^\text{\textbf{2}}$, based on graph attention network~\cite{velivckovic2017graph}.
Its basic version, Net$^\text{\textbf{2f}}$, intends to be fast yet effective. 
The other version, which emphasizes more on accuracy and is
denoted as Net$^\text{\textbf{2a}}$, 
captures rich global information with a highly flexible model
through circuit partitioning. 

Recently, deep learning has generated a huge impact on many applications where data is represented in Euclidean space. However, there is a wide range of applications where data is in the form of graphs. Machine learning on graphs is much more challenging as there is no fixed neighborhood structure like in images. All neural network-based methods on graphs are referred to as graph neural networks (GNN). The most widely-used GNN methods include graph convolution network (GCN) \cite{kipf2016semi}, graphSage (GSage) \cite{hamilton2017inductive}, and graph attention network (GAT) \cite{velivckovic2017graph}. They all convolve each node's representation with its neighbors' representations, to derive an updated representation for the central node. Such operation essentially propagates node information along edges and thereby topology pattern is learned.

Similarly, in Electronic Design Automation (EDA), circuit designs are embedded in Euclidean space after placement, which inspired many convolutional neural network (CNN)-based methods~\cite{xie2018routenet,fang2018machine,xie2020powernet}. But before placement, a circuit structure is described as a graph and spatial information is not yet available. Till recent years, GNN is explored for EDA applications~\cite{ma2019high,zhang2019circuit}.  
The work in \cite{ma2019high} predicts observation point candidates with a model similar to GSage~\cite{hamilton2017inductive}. Graph-CNN \cite{zhang2019circuit} predicts the electromagnetic properties of post-placement circuits. This method is limited to very small-scale circuit graphs with less than ten nodes. Overall, GNN has great potential but is much less studied than CNN in EDA.

In this work, our contributions include:

\begin{itemize}
\item As far as we know, this is the first work making use of GNN for pre-placement net length estimation. GNNs (GCN, GSage, GAT) outperform both conventional heuristics and common machine learning methods in almost all measured metrics. 
\item We propose to extract global topology information through partitioning. Based on partition results, we define innovative directional edge features between nets, 
which substantially contribute to Net$^\text{\textbf{2}}$'s superior accuracy.
\item We propose a GAT-based model named Net$^\text{\textbf{2}}$, which is customized for this net length problem. It includes a fast version Net$^\text{\textbf{2f}}$, which is 1000$\times$ faster than placement, 
and an accuracy-centric version Net$^\text{\textbf{2a}}$, which effectively extracts global topology information from unseen netlists and significantly outperforms plug-in use of existing GNN techniques.
\item To focus on nets, we propose a graph construction method that treats nets as nodes. In designing Net$^\text{\textbf{2}}$ architecture, we define different convolution layers for graph nodes and edges to incorporate both edge and node features.
\item Besides net length, we further apply the proposed methods to estimate path length. Net$^\text{\textbf{2}}$ also proves to be superior in identifying long paths.
\end{itemize}

%\item For the features of WireGraph, we define novel edge features based on efficient clustering algorithms. They prove to strongly correlate with distance between cells after placement.

\section{Problem Formulation}

\begin{figure}[!t]
  \centering
    \includegraphics[width=0.9\columnwidth]{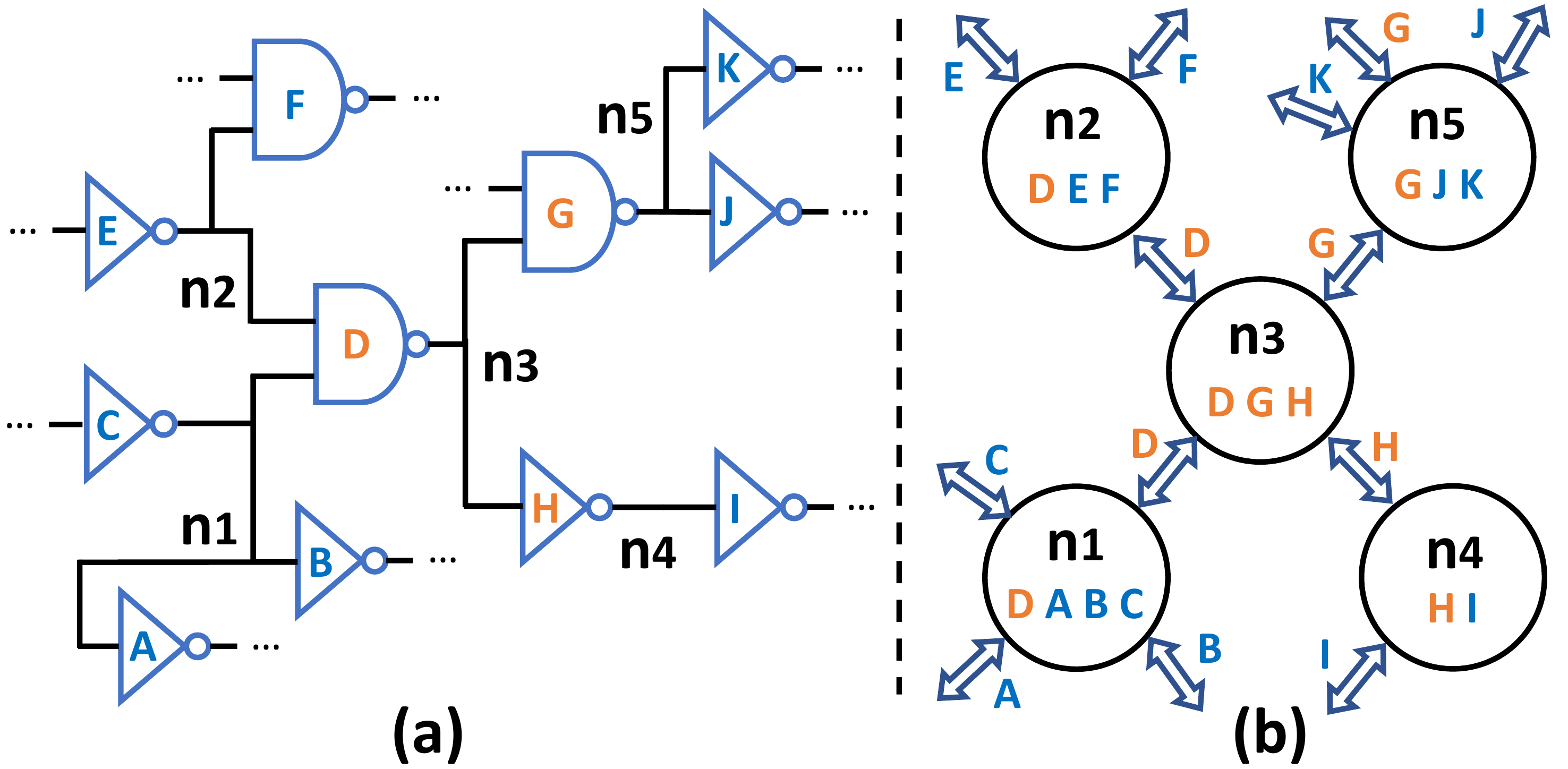}
        \vspace{-3mm}
  \caption{(a) Part of a netlist. (b) The corresponding graph.}
  \label{method_circuit}
\end{figure}

We define terminologies with Figure~\ref{method_circuit}. Figure~\ref{method_circuit}(a) shows part of a netlist, including five nets \{$n_1$, $n_2$, $n_3$, $n_4$, $n_5$\} and 11 cells \{$c_A$, $c_B$, ..., $c_K$\}. Now we focus on net $n_3$, which touches 3 cells \{$c_D$, $c_G$, $c_H$\} and is referred to as a \emph{3-pin net}. Its \emph{driver} is cell $c_D$; its \emph{sinks} are cells \{$c_G$, $c_H$\}. We denote the area of $n_3$'s driver cell as $a_{dri}^3$. Net $n_3$'s \emph{fan-ins} $N^3_{in} =$ \{$n_1$, $n_2$\} ; its \emph{fan-outs} $N^3_{out} =$ \{$n_4$, $n_5$\}. Its \emph{fan-in size} is 2, denoted as $|N^3_{in}| = 2$. Its \emph{fan-out size} (number of sinks) is 2, denoted as $|N^3_{out}| = 2$. Every net can have only one driver but multiple sinks. Thus, the number of cells $= 1 + |N^3_{out}| = 3$ for this net. Net $n_3$'s one-hop neighbors include both its fan-in and fan-out: $\mathcal{N}(n_3) = N_{in}^3 \cup N_{out}^3 = $ \{$n_1$, $n_2$, $n_4$, $n_5$\}. The number of its neighbors is also known as the degree of $n_3$: $deg(n_3) = |\mathcal{N}(n_3)| = 4$.

\begin{table}[!t]
  \centering
  \caption{Commonly Used Notations}
  \vspace{-3mm}
  \label{tbl:notation}
  \resizebox{\linewidth}{!}{
  \begin{tabular}{l | l  | l | l}
 	\hline
 	\multirow{2}{*}{Notation}  &  \multirow{2}{*}{Description} &  \multirow{2}{*}{Notaion} &  \
 	 \multirow{2}{*}{Description}    \\
 	  &     &     &          \\
	\hline
	$n_k$             & net or node    &  $O_k$  & node features     \\
	\hline
	\{$c_G$, $c_H$\}  & cells          &  $E_{b \rightarrow k}$  & edge features  \\
	\hline
	$a^k_{dri}$ &   driver's area              &  $P[c_H]$    &  cell partition IDs \\
	\hline
	$N^k_{in}$, $N^k_{out}$  &  fan-in/fan-out nets         &  $M[n_k]$   &  node partition IDs \\	
	\hline
	$\mathcal{N}(n_k)$ & 1-hop neighbors       &   $h^{(t-1)}_k$, $h^{(t)}_k$ & GNN embeddings  \\
	\hline
	$|N^k_{in}|$, $|N^k_{out}|$      &  fan-in/fan-out size     &   $W^{(t)}$, $\theta^{(t)}$ & learnable weights  \\
	\hline
	$deg(n_k)$       &  degree of net                       &  $g(\,), \sigma(\,)$ & activation functions  \\
	\hline
	$L^k$             &  net length label                   &  $s(\,)$, $\mu(\,)$ & std deviation, mean \\
	\hline
	$n_b \rightarrow n_k$ & directional edge                &  $[\,\,\,||\,\,\,]$ & concat to one list \\
	\hline
	$c_{kb}$ or $c_{bk}$ &   cell on $n_b \rightarrow n_k$   &  $\backslash$ & exclude from list \\ 
	\hline
  \end{tabular}
  }
  %\vspace{-2mm}
\end{table}

The net length is the label for training and prediction. Each net's length is the half perimeter wirelength (HPWL) of the bounding box of the net after placement. The length of net $n_k$ is $L^k$. 

To apply graph-based methods, we convert each netlist to one directed graph. Different from most GNN-based EDA tasks, net length prediction focuses on nets rather than cells. Thus we represent each net as a node, and use the terms \emph{node} and \emph{net} interchangeably. For each net $n_k$, it is connected with its fan-ins and fan-outs through their common cells by edges in both directions. The common cell shared by both nets on that edge is called its \emph{edge cell}. For example, in Figure \ref{method_circuit}(b), net $n_3$ is connected with nets $n_4$ and $n_5$ through its sinks $c_G$ and $c_H$; it is connected with nets $n_1$ and $n_2$ through its driver $c_D$. The edges through edge cell $c_G$ is denoted as $n_3 \rightarrow n_5$ and $n_5 \rightarrow n_3$. The edge cell $c_G$ can also be referred to as $c_{35}$ or $c_{53}$. We differentiate edges in different directions because we will assign different edge features to $n_3 \rightarrow n_5$ and $n_5 \rightarrow n_3$. 

An important concept throughout this paper is global and local topology information. We use the number of hops to denote the shortest graph distance between two nodes on a graph. The \emph{information} of each net refers to its number of cells and driver's area. Local information includes the information about the estimated net itself, or from its one to two-hop neighboring nets. In contrast, global information means the pattern behind the topology of the whole netlist or the information from nets far away from the estimated net $n_k$. The range of neighbors that can be accessed by each model is referred to as the model's \emph{receptive field}.

\section{Challenges}
\label{Challenges}

\begin{figure}[!htb]
  \centering
  \vspace{-3mm}
    \includegraphics[width=0.9\columnwidth]{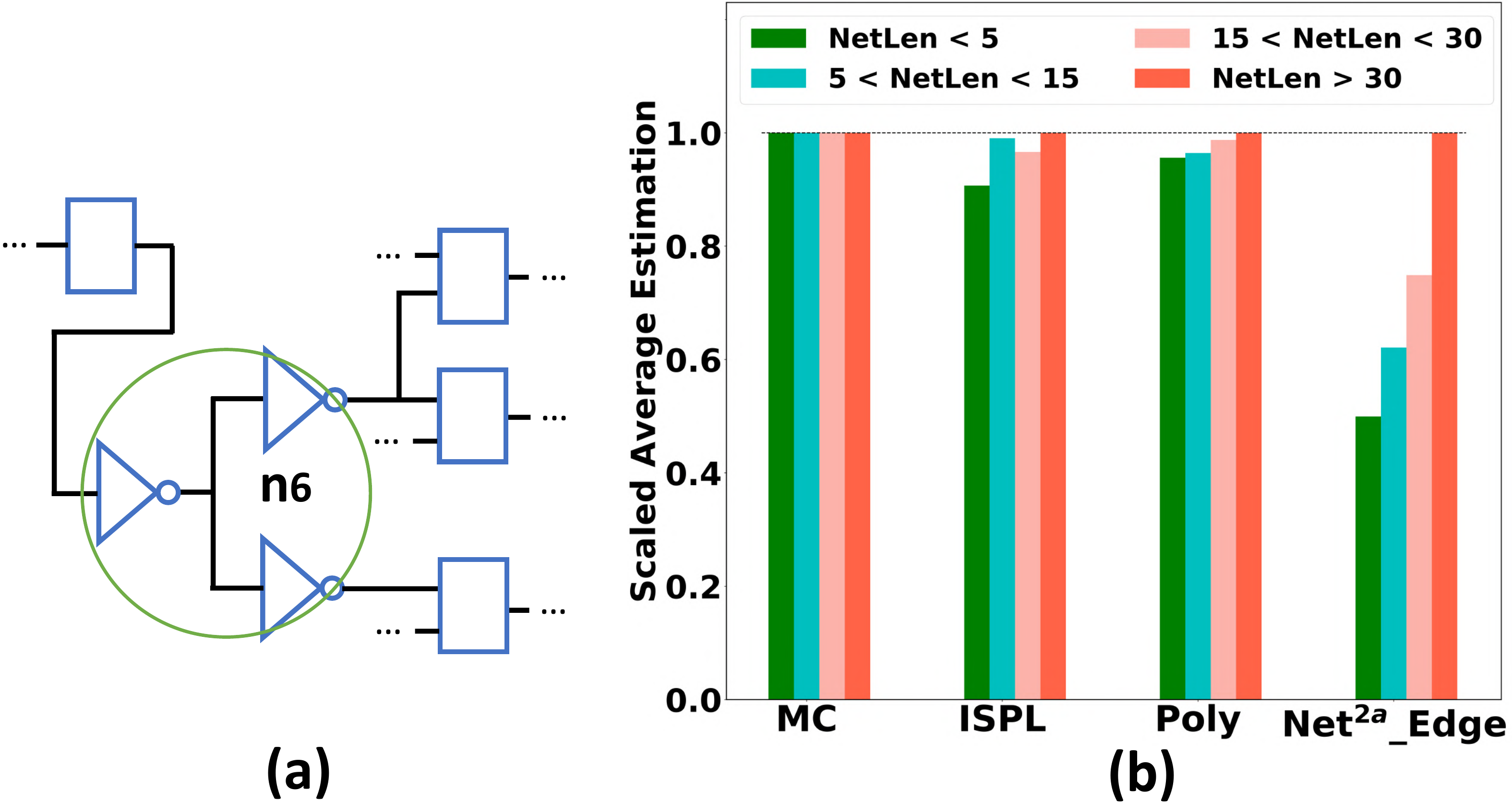}
    \vspace{-3mm}
  \caption{(a) A typical local topology. (b) Predictions on nets with similar local topology information.}
  \label{example}
\end{figure}

We provide an example to show the challenge in net length prediction and the importance of global information. Figure \ref{example}(a) shows a net $n_6$ with a commonly seen local topology information: $|N^6_{in}| = 1$, $|N^6_{out}|= 2$, $deg(n_6) = 3$. For its neighbors, it has one 2-pin fan-in, one 2-pin fan-out and one 3-pin fan-out.

In a netlist of design B20 in ITC 99 \cite{corno2000rt}, we find 725 nets with exactly the same driver cell's area, number of cells and one-hop neighbor information as $n_6$, but their net lengths after placement range from 1 \si{\um} to more than 100 \si{\um}. Distinguishing these similar nets is highly challenging without rich global information. To prove this, Figure \ref{example}(b) shows the prediction from different methods on these 725 similar nets. These nets are firstly divided into four different types according to their actual net length, each type with $432, 190, 67, 36$ nets, respectively. We then plot the scaled averaged estimation by different methods for each type of net. MC \cite{hu2003wire}, which only looks at one-hop neighbors, cannot distinguish these nets at all. ISPL \cite{kahng2005intrinsic}, which captures some global information by searching shortest path, gives a slightly lower estimation on the shortest type (netLen < 5 \si{\um}). By looking at two-hop neighbors, a polynomial model with pre-defined features (Poly) \cite{bodapati2001prelayout} \cite{fathi2009pre} captures the trend with a tiny difference between different types. For Net$^\text{\textbf{2}}$, we only train its edge convolution layer on other designs and present its output. Its estimations on different net types differ significantly. This example shows the importance of global information in distinguishing a large number of nets with similar local information.

\section{Algorithm}

\subsection{Node Features on Graph}

\begin{algorithm}[!b]
\caption{Graph Generation with Node Features}
\label{alg1}
\begin{flushleft}
  \textbf{Input}: Basic features \{$|N_{in}^k|$, $|N_{out}^k|$, $a_{dri}^k$\}, net length label $L^k$, the fan-in nets $N_{in}^k$ and fan-out nets $N_{out}^k$ of each net $n_k$. \\
  \textbf{Generate Node Features}:
\end{flushleft}
\begin{algorithmic}[1]
\For{each net $n_k$}
\State $a_{all} = [a_{dri}^k]$, $in_{in} = []$, $in_{out} = []$, $out_{in}= []$, $out_{out} = []$
\For{each net $n_i \in N_{in}^k$, each net $n_o \in N_{out}^k$} \label{line:it_i}
%\State $a_{all}$ .add ($a_{dri}^o$) \label{line:it_oa}
%\State $in_{in}$  .add ($|N_{in}^i|$)
%\State $in_{out}$ .add ($|N_{out}^i|$)
\BiState{$in_{in}$.add ($|N_{in}^i|$) \textbf{;}}{$in_{out}$.add ($|N_{out}^i|$)}
\BiState{$out_{in}$.add ($|N_{in}^o|$) \textbf{;}}{$out_{out}$.add ($|N_{out}^o|$)}
\State $a_{all}$.add($a_{dri}^o$) \label{line:it_oa}
%\State $out_{in}$  .add ($|N_{in}^o|$)
%\State $out_{out}$ .add ($|N_{out}^o|$)
\EndFor
\State $O_k$ = \{$|N_{in}^k$|, $|N_{out}^k|$, $a_{dri}^k$, $\sum {a_{all}}$, $\sum{out_{in}}$, $\sum{out_{out}}$, \newline 
\hspace*{12mm} $\sum in_{in}, \sum in_{out}$, $s(out_{in})$, $s(out_{out})$, $s(in_{in})$, $s(in_{out})$\}\label{line:node_f}
\EndFor
\end{algorithmic} 
\begin{flushleft}
\textbf{Build Graph}:
\end{flushleft}
\begin{algorithmic}[1]
\State Initiate a graph $G$. Each net is a node.
\For{each net $n_k$}
\State For node $n_k$ in $G$, set $O_k$ as node feature, $L^k$ as label.
\For{each net $n_b \in N_{in}^k \cup N_{out}^k$}
\State Add directed edge $n_b$ \textrightarrow $n_k$.
\EndFor
\EndFor
\end{algorithmic} 
\begin{flushleft}
  \textbf{Output}: Graph $G$ with node features $O$ and label $L$.
\end{flushleft}
\end{algorithm}

Algorithm \ref{alg1} shows how we build a directed graph and generate features for each node with a given netlist.
On average, a net with more large cells tends to be longer. Thus, the most basic net features include the net's driver's area, fan-in and fan-out size \{$|N^k_{in}|$, $|N^k_{out}|$, $a_{dri}^k$\}. Feature $\sum a_{all}$ is the sum of areas over all cells in $n_k$. It is calculated by including the drivers of all $n_k$'s fan-outs in line \ref{line:it_oa}, which are the sinks of $n_k$. Besides these basic features, we capture the more complex impact from neighbors. As shown in lines \ref{line:it_i}, we go through all neighbors of $n_k$ to collect their fan-in and fan-out sizes. The summation $\sum$ and standard deviation $s()$ of these neighboring information are added to node features $O_k$ in line \ref{line:node_f}.

%%%%%%%%%%%%%%%%%%%%%%%%%%%%%%%% subsection 4.2

\subsection{Edge Features}

\begin{algorithm}[!t]
\caption{Define Edge Features on Graph}
\begin{flushleft}
  \textbf{Input}: Cell cluster ID $P[c_k]$ for each cell $c_k$, net cluster ID $M[n_k]$ and the neighbors $\mathcal{N}(n_k)$ of each net $n_k$. Directed graph $G$. \\
\end{flushleft}
\label{alg2}
\begin{algorithmic}[1]
\Function{measureDiff}{$c_{bk}$, $n_b$, $c_{ok}$, $n_o$} 
%\hspace*{3mm} // estimates distance between $c_{bk}$ and $c_{ok}$
    \State $f_0 = 1 - (P[c_{bk}] == P[c_{ok}])$  \label{line:f0}
    \State $P_b = [P[c] $ for $ c \in n_b] $ // cluster IDs for $n_b$'s cells \label{line:f1s}
    \State $P_o = [P[c] $ for $ c \in n_o] $ // cluster IDs for $n_o$'s cells
    \State $P_{b\_not\_o} = P_b \backslash P_o$ $\qquad$ // IDs in $P_b$ but not in $P_o$
    \State $P_{o\_not\_b} = P_o \backslash P_b$ $\qquad$ // IDs in $P_o$ but not in $P_b$
    \State $f_1 = \frac{|P_{b\_not\_o}|}{|P_b|} + \frac{|P_{o\_not\_b}|}{|P_o|}$ // percent of different IDs \label{line:f1e}
    \State $f_2 = 1 - (M[n_b] == M[n_o])$
    \State return [$f_0$, $f_1$, $f_2$]
\EndFunction
\State
\For{each net $n_k$}
\For{each net $n_b \in \mathcal{N}(n_k)$}
\State $F_0$ = [], $F_1$ = [], $F_2$ = []
\State Cell $c_{bk}$ is the edge cell on $n_b$ \textrightarrow $n_k$
\State Other neighbors $N_{other}^k = \mathcal{N}(n_k) \backslash \{n_b\}$
\For{each net $n_o \in N_{other}^k$}
\State Cell $c_{ok}$ is the edge cell on $n_o$ \textrightarrow $n_k$
\State $f_0$, $f_1$, $f_2$ = \textproc{measureDiff} ($c_{bk}$, $n_b$, $c_{ok}$, $n_o$)
%\State $F_0$ .add ($f_0$)
%\State $F_1$ .add ($f_1$)
%\State $F_2$ .add ($f_2$)
\State $F_0$.add($f_0$) \textbf{;} \ \ $F_1$.add($f_1$) \textbf{;} \ \  $F_2$.add($f_2$)
\EndFor
\State  $f_3 = 1 - (M[n_b] == M[n_k])$
\State $E_{b \rightarrow k}$ = \{$\sum F_0$, $\mu (F_0)$, $\sum F_1$, $\mu (F_1)$, $\sum F_2$, $\mu (F_2)$, $f_3$\}\label{line:edge_comb}
\State Set $E_{b \rightarrow k}$ as the feature of edge $n_b \rightarrow n_k$ in $G$.
\EndFor
\EndFor
\end{algorithmic} 
\begin{flushleft}
  \textbf{Output}: Graph $G$ with edge features $E$.
\end{flushleft}
\end{algorithm}

In Algorithm \ref{alg1}, node features $O_k$ include up to two-hop neighboring information. The receptive field of the GNN method itself depends on the model depth, which is usually two to three layers. Thus the model can reach as far as four to five-hop neighbors, which is already more than previous works. 
This work goes beyond that to capture more global information from the whole graph.

To capture global information, we use an efficient multi-level partitioning method hMETIS \cite{karypis1999multilevel} to divide one netlist into multiple clusters / partitions. The partition method minimizes the overall cut between all clusters, which provides a global perspective. %We then define multiple novel edge features based on them. 
We denote the partition result as $M$. Each net $n_k$ is assigned a cluster ID $M[n_k]$, which denotes the cluster it belongs to. To generate more information, on the same netlist, we also build a hyper-graph $HG_c$ by using cells as nodes. In $HG_c$, each hyper-edge corresponds to a net. Similarly, the partition result on $HG_c$ is denoted as $P$. Each cell $c_k$ is assigned a cluster ID $P[c_k]$. Notice that $HG_c$ is only used to generate cluster ID for each cell. 

Cluster IDs are not directly useful by themselves. What matters in this context is the difference in cluster IDs between cells and nets. Algorithm \ref{alg2} shows how the cluster information is incorporated into GNN models through novel edge features $F_0, F_1, F_2, f_3$. The most important intuition behind this is: for a high-quality placement solution, on average, the cells assigned to different clusters tend to be placed far away from each other.

In Algorithm \ref{alg2}, we design the edge features to quantify the source node's contribution to the target node's length. The contribution here means the source net is ``pulling'' the edge cell far away from other cells in the target net. The edge features measure such ``pulling'' strength. When the edge cell is ``pulled'' away, the target net results in a longer length. In Algorithm \ref{alg2}, for edge $n_b$ \textrightarrow $n_k$, function \textproc{measureDiff} measures the difference in assigned clusters between node $n_b$ and every other neighboring node $n_o$, which indicates the distance between $c_{bk}$ and $c_{ok}$. If the distance between edge cell $c_{bk}$ and every other cell $c_{ok}$ in $n_{k}$ is large, it means $c_{bk}$ is placed far away from other cells in net $n_k$. In this case, edges features $F_0, F_1, F_2, f_3$ are large. That is why edge features imply how strong the edge cell is ``pulled'' away from the target node.

\begin{figure}[!tb]
  \centering
  \vspace{-3mm}
    \includegraphics[width=\columnwidth]{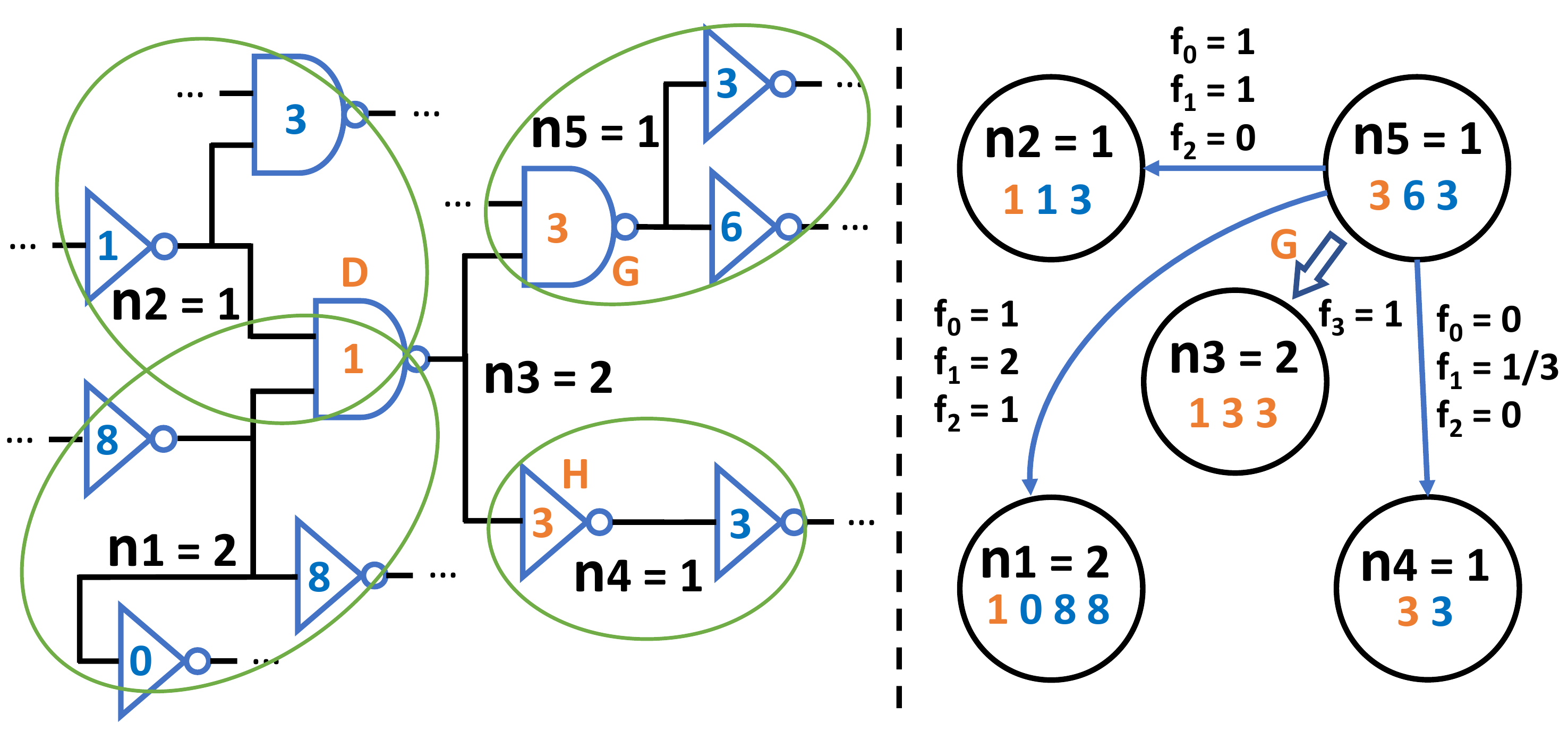}
    	\vspace{-3mm}
  \caption{Define edge features by partition results.}
  \label{method_cluster}
\end{figure}

Figure \ref{method_cluster} shows an example of Algorithm \ref{alg2} using the netlist same as Figure \ref{method_circuit}. The number on each cell or net is the cluster ID assigned to it after partition. Figure \ref{method_cluster} measures the edge features of edge $n_5$ \textrightarrow $n_3$, representing how strongly edge cell $c_G$ is pulled by $n_5$ from both cells \{$c_D$, $c_H$\} in $n_3$. To calculate this, we measure the distance between $c_G$ and $c_H$ by \textproc{measureDiff}($c_G$, $n_5$, $c_H$, $n_4$) in Algorithm \ref{alg2}; and the distance between $c_G$ and $c_D$ by \textproc{measureDiff}($c_G$, $n_5$, $c_D$, $n_1$) and \textproc{measureDiff}($c_G$, $n_5$, $c_D$, $n_2$).

Take \textproc{measureDiff}($c_G$, $n_5$, $c_H$, $n_4$) as an example to show how it measures distance between $c_G$ and $c_H$. As shown in line \ref{line:f0}, feature $f_0$ measures the difference in $c_G$ and $c_H$' cluster IDs, $f_0 = 1 - (P[c_G] == P[c_H]) = 1 - (3 == 3) = 0$. Feature $f_1$ measures the difference in all cells between $n_5$ and $n_4$. As shown from line \ref{line:f1s} to \ref{line:f1e}, $P_5 = [3, 6, 3]$ and $P_4 = [3, 3]$. Then $P_{5\_not\_4} = [6]$ and $P_{4\_not\_5} = []$. They are normalized by the number of cells $|P_5| = 3$ and $|P_4| = 2$, in order to avoid bias toward nets with many cells. Thus, $f_1 = \frac{1}{3} + \frac{0}{2} = \frac{1}{3}$. Feature $f_2$ measures the difference between $n_5$ and $n_4$, $f_2 = 1- (M[n_5] == M[n_4]) = 1 - (1==1) = 0$. As this example shows, we only measure whether cells / nets have the same cluster IDs, and the order of IDs does not matter.

After measuring the difference in cluster ID between $c_{G}$ and all other cells in $n_3$, for the edge $n_5$ \textrightarrow $n_3$, $F_0 = [1, 1, 0]$; $F_1 = [2, 1, \frac{1}{3}]$; $F_2 = [1, 0, 0]$. $f_3$ measures the difference between $n_5$ and $n_3$, $f_3=1$. This example shows how we incorporate global information from partition into edge features. Actually, we generate multiple different partitioning results $M$, $P$ by requesting different number of clusters. That results in multiple different \{$F_0$, $F_1$, $F_2$, $f_3$\}. All these different edge features are processed in line \ref{line:edge_comb} and concatenated together as the final edge features $E_{b \rightarrow k}$.

\subsection{Common GNN Models}

\setlength{\belowdisplayskip}{3pt} \setlength{\belowdisplayshortskip}{3pt}
\setlength{\abovedisplayskip}{-5pt} \setlength{\abovedisplayshortskip}{-5pt}

This section introduces how GNN models are applied on the graph G we build. GNN models are comprised of multiple sequential convolution layers. Each layer generates a new embedding for every node based on the previous embeddings. For node $n_k$ with node features $O_k$, denote its embedding at the $t^{\text{th}}$ layer as $h_k^{(t)}$. Its initial embedding is the node features $h_k^{(0)} = O_k$. Sometimes the operation includes both neighbours and the node itself, we use $n_{\beta}$ to denote it: $n_{\beta} \in \mathcal{N}(n_k) \cup \{n_k\}$. In each layer $t$, GNNs calculate the updated embedding $h_k^{(t)}$ based on the previous embedding of the node itself $h_k^{(t-1)}$ and its neighbors $h_b^{(t-1)} | n_b\in \mathcal{N}(n_k)$.

We show one layer of GCN, GSage, and GAT below. Notice that there exists other expressions. The two-dimensional learnable weight at layer $t$ is $W^{(t)}$. In GAT, there is an extra one-dimensional weight $\theta^{(t)}$. The operation $[\: ||\: ]$  concatenates two vectors into one longer vector. Functions $\sigma$ and $g$ are sigmoid and Leaky ReLu activation function, respectively.

On GCN (with self-loops), $\mathcal{F}_{GCN}^{(t)}$ \cite{kipf2016semi} is:

\begin{gather*}
  h_k^{(t)} = \sigma (\sum_{n_{\beta}\in {\mathcal{N}(n_k) \cup \{n_k\}}} a_{k\beta} W^{(t)} h_\beta^{(t-1)}) \\
  \text{where   } a_{k\beta} = \frac{1}{\sqrt{deg(k) + 1} \sqrt{deg(\beta) + 1}} \in \mathbb{R}
\end{gather*}

On GSage, $\mathcal{F}_{GSage}^{(t)}$ \cite{hamilton2017inductive} is:
\vspace{3pt}
\begin{gather*}
  h_k^{(t)} = \sigma ( W^{(t)} [h_k^{(t-1)} || \frac{1}{deg(k)} \sum_{n_b\in {\mathcal{N}(n_k)}}  h_b^{(t-1)}]) 
\end{gather*}

On GAT, $\mathcal{F}_{GAT}^{(t)}$ \cite{velivckovic2017graph} is:
\vspace{5pt}
\begin{gather*}
  h_k^{(t)} = \sigma (\sum_{n_\beta\in {\mathcal{N}(n_k) \cup \{n_k\}}} a_{k\beta} W^{(t)} h_\beta^{(t-1)}) \qquad \qquad \\
\end{gather*}
\vspace{-5mm}
\begin{align*}
  \text{where   } a_{k\beta} &= softmax_\beta (r_{k\beta})  \text{ \,\; over $n_k$ and its neighbors,} \\
  r_{k\beta} &= g (\theta^{(t)\intercal} [ W^{(t)} h_\beta^{(t-1)} || W^{(t)} h_k^{(t-1)} ]) \in \mathbb{R}
\end{align*}

Here we briefly discuss the difference between these methods. GCN scales the contribution of neighbors by a pre-determined coefficient $a_{k\beta}$,  depending on the node degree. In contrast, GAT uses learnable weights $W$, $\theta$ to firstly decide node $n_\beta$'s contribution $r_{k\beta}$, then normalize the coefficient $r_{k\beta}$ across $n_k$ and its neighbors through a softmax operation. Such a learnable $a_{k\beta}$ %for each node
leads to a more flexible model. GSage does not scale neighbors by any factor. For all these GNN methods, the last layer 's output embedding $h_k^{(t)}$ is connected to a multi-layer ANN.

\subsection{Net$^\text{\textbf{2}}$ Model}

The node convolution layer of the Net$^\text{\textbf{2}}$ is based on GAT, considering its higher flexibility in deciding neighbors' contribution $a_{k\beta}$. Thus node convolution layer is $\mathcal{F}_{GAT}^{(t)}$. In the final embedding, we concatenate the outputs from all layers, instead of only using the output of the final layer like most GNN works. This is a customization, by which the embedding includes contents from different depths. The shallower one includes more local information, while the deeper one contains more global information. Such an embedding provides more information for the ANN model at the end and may lead to better convergence. The idea of combining shallow and deep layers has inspired many classical deep learning methods in Euclidian space \cite{he2016deep} \cite{ronneberger2015u}, but it is not widely applied in GNNs for node embeddings. After three layers of node convolution, the final embedding for each node is $[h_k^{(1)} || h_k^{(2)} || h_k^{(3)}]$. Without partitioning, this is the embedding for our fast solution Net$^\text{\textbf{2f}}$.

In order to utilize edge features, here we define our own edge convolution layers $\mathcal{E}$ as customization. For each directed edge $n_b$ \textrightarrow $n_k$, we concatenate both target and source nodes' features $[O_k || O_b]$ together with its edge features $E_{b\rightarrow k}$ as the input of edge convolution. Combining node features when processing edge features enables $\mathcal{E}$ to distinguish different edges with similar edge features. The output embedding is:

\begin{gather*}
e_{k\_sum} = \sum_{n_b \in \mathcal{N}(n_k)} W_2 W_1 [O_k || E_{b\rightarrow k} || O_b] \\
e_{k\_mean} = \frac{1}{deg(k) }e_{k\_sum}
\end{gather*}

The two two-dimensional learnable weights $W_1$ and $W_2$ can be viewed as applying a two-layer ANN to the concatenated input. We choose two-layer ANN rather than one-layer here because the input vector $[O_k || E_{b\rightarrow k} || O_b]$ is long and contains heterogeneous information from both edge and node. We prefer to learn from them with a slightly more complex function. After the operation, both $e_{k\_sum}$ and $e_{k\_mean}$ are on nodes. Then, we add an extra node convolution using the output from edge convolution as input. This structure learns from neighbors' edge embeddings $e_{b\_sum}, e_{b\_mean}$.

\begin{equation*}
h_k^{(e)} = \mathcal{F}_{GAT}^{(e)} ([e_{k\_sum} || e_{k\_mean}], [e_{b\_sum} || e_{b\_mean}])
%h_k^{(e)} = \mathcal{F}_{GAT}^{(e)} ([e_{k\_sum} || e_{k\_mean}])
\end{equation*}

%Thus, for our accurate solution Net$^\text{\textbf{2a}}$, its final embedding for each node is $[h_k^{(1)} || h_k^{(2)} || h_k^{(3)} || e_{k\_sum} || e_{k\_mean} || h_k^e]$. For both Net$^\text{\textbf{2f}}$ and Net$^\text{\textbf{2a}}$, their final embeddings are then connected to an ANN.

Inspired by the same idea in Net$^\text{\textbf{2f}}$, we combine the contents from all layers for our accurate solution Net$^\text{\textbf{2a}}$. Its final embedding is $[h_k^{(1)} || h_k^{(2)} || h_k^{(3)} || e_{k\_sum} || e_{k\_mean} || h_k^{(e)}]$. For both Net$^\text{\textbf{2f}}$ and Net$^\text{\textbf{2a}}$, their final embeddings are then connected to an ANN.

\section{Experimental Results}

\subsection{Experimental Setup}

\begin{table}[!t]
  \vspace{-1mm}
  \centering
  \caption{ Number of Cells in Netlists for Each Design}
  \vspace{-3mm}
  \label{tbl:designs}
  \resizebox{0.97\linewidth}{!}{
  \begin{tabular}{l | c c  c c c  c c}
 	\hline
 	\multirow{2}{*}{}  &  \multirow{2}{*}{B14} &  \multirow{2}{*}{B15} &  \
 	 \multirow{2}{*}{B17} &  \multirow{2}{*}{B18} &   \multirow{2}{*}{B20} &  \
 	  \multirow{2}{*}{B21} &   \multirow{2}{*}{B22}  \\
 	  \\
	\hline
	Smallest  &  13$\,$K  & 5.3$\,$K  &  18$\,$K  &  54$\,$K &  26$\,$K  &  26$\,$K  & 39$\,$K \\
	Largest   &  34$\,$K  & 15$\,$K   &  49$\,$K  & 138$\,$K &  67$\,$K  &  66$\,$K  & 99$\,$K \\
	\hline
  \end{tabular}
  }
\end{table}

Seven different designs from ITC99 \cite{corno2000rt} are synthesized with Synopsys Design Compilier\textregistered \  in 45nm NanGate Library, and then placed by Cadence Innovus\texttrademark \ v17.0. For each design, we synthesize 10 different netlists with different synthesis parameters. Altogether there are 70 different netlists and corresponding placement solutions. Table \ref{tbl:designs} shows the size of these netlists. When testing ML models on each design, the model is only trained on the 60 netlists from the \emph{other} six designs to prevent information leakage. Then the result is averaged over the 10 netlists of the tested design.

All GNNs are built with Pytorch 1.5 \cite{paszke2019pytorch} and Pytorch-geometric \cite{Fey/Lenssen/2019}. The partition on graphs is performed by hMETIS \cite{karypis1999multilevel} executable files. The experiment is performed on a machine with a Xeon E5 processor and an Nvidia GTX 1080 graphics card.

Here we introduce the best hyper-parameters after parameter tuning.
For all GNN methods, we use three layers of GNN with two layers ANN. The attention head number of GAT is two. The size of each node convolution output is 64. The size of edge convolution output is twice of the input size $[O_k||E_{b \rightarrow k}||O_b]$. The size of the first-layer ANN is the same as its input embedding, and the size of the second-layer ANN is 64. A batch normalization layer is applied after each GNN layer for better convergence. Because of the difference in graph size, each batch includes only one graph, and the training data is shuffled during training. We use stochastic gradient descent (SGD) with learning rate 0.002 and momentum factor 0.9 for optimization. GNN models converge in 250 epoches.

When partitioning each netlist, we generate seven different cell-based partitions $P$ by requesting the number of output clusters to be the number of cells divided by 100, 200, 300, 500, 1000, 2000, and 3000. Because different partitions are generated in parallel, the overall runtime depends on the slowest one. Similarly, we generate three net-based partitions $M$ by requesting the cluster number to be the number of nets divided by 500, 1000, and 2000. These cluster numbers are achieved by tuning during experiments, which provides good enough coverage over different cluster sizes.
 
Representative previous methods MC~\cite{hu2003wire}, ISPL \cite{kahng2005intrinsic}, and Poly~\cite{fathi2009pre} are implemented for comparisons. 
As for traditional ML models, we implement a three-layer ANN model using node features $O$.

Here we summarize the receptive field of all methods. MC is limited to one-hop neighbors, while Poly and ANN can reach two-hop neighbors. The receptive field of ISPL varies among different nodes. According to \cite{kahng2005intrinsic}, ISPL for most nets is within several hops. In comparison, all three-layer GNNs and Net$^\text{\textbf{2f}}$ can access five-hop neighbors. Net$^\text{\textbf{2a}}$ measures the impact from the whole netlist.

\begin{figure}[!t]
  \centering
    \includegraphics[width=0.9\columnwidth]{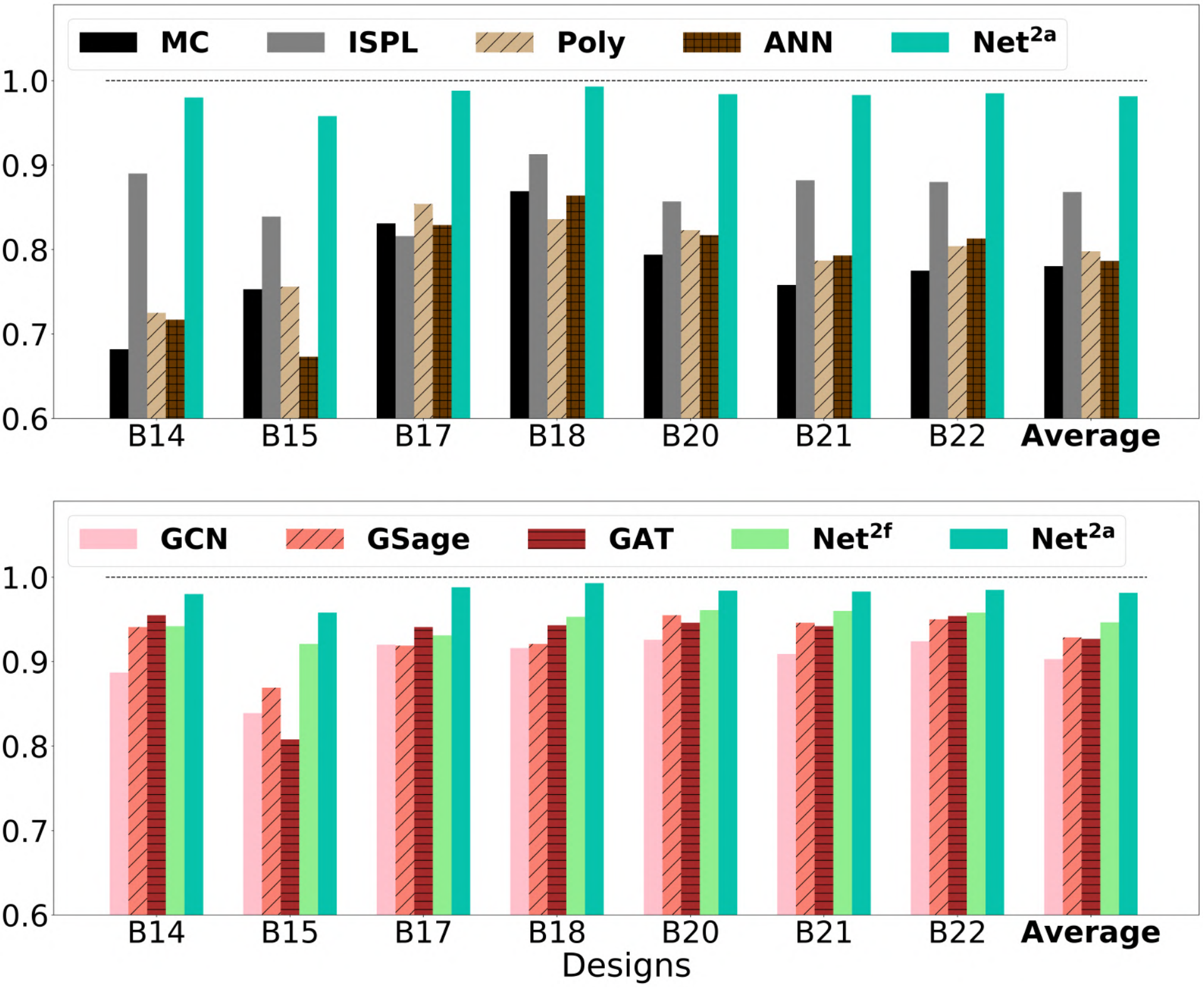}
    \vspace{-3mm}
  \caption{Correlation coefficient R between prediction and label measured in 20 bins. Net$^\text{\textbf{2a}}$ is reported in both subplots.}
  \label{bin_r}
\end{figure}

\subsection{Correlation on Net Length}

To measure the correlation between prediction and ground truth, we apply a classical criterion used in many net length estimation works~\cite{fathi2009pre,kahng2005intrinsic,liu2004pre}. Firstly, we calculate a range of net length labels [$L_{0\%min}$, $L_{95\%max}$]. It means from the shortest net length (close to zero) to the 95 percentile largest net length. The top 5\% longest nets are excluded to prevent an extraordinarily large range. Then, the calculated range is partitioned into 20 equal bins and the average of both predictions and labels in each bin are calculated. Figure \ref{bin_r} shows the correlation coefficient between these 20 averaged predictions and labels. Unlike some previous works, we define such 20 bins using labels instead of predictions for a fair comparison. In this way, the nets assigned to each bin are kept the same for different methods. On average, the correlation coefficient for GraphSage and GAT approaches 0.93. Net$^\text{\textbf{2f}}$ further improves it to around 0.95. In comparison, the best traditional method is ISPL, with correlation equals 0.87. The correlation for Net$^\text{\textbf{2a}}$ is higher than 0.98.

\begin{table}[!b]
  \centering
  \vspace{1mm}
  \caption{Identify 10\% Longest Nets in ROC AUC (\%)}
    \vspace{-3mm}
  \label{tbl:10LongestNet}
  \resizebox{0.97\linewidth}{!}{
  \begin{tabular}{l | c c c c c c c | c}
 	\hline
 	\multirow{2}{*}{Methods}  &  \multirow{2}{*}{B14} &  \multirow{2}{*}{B15} &  \
 	 \multirow{2}{*}{B17} &  \multirow{2}{*}{B18} &   \multirow{2}{*}{B20} &  \
 	  \multirow{2}{*}{B21} &   \multirow{2}{*}{B22} &  \multirow{2}{*}{Ave} \\
 	  &     &     &     &     &     &     &     &     \\
	\hline
	NumCell  &  71.5  &  67.5  &  65.7  &  68.8  &  71.5  &  71.7  &  71.8  &  69.8   \\
	MC       &  73.1  &  70.3  &  71.3  &  72.8  &  75.0  &  74.3  &  74.8  &  73.1   \\
	ISPL     &  79.9  &  71.9  &  69.2  &  71.1  &  78.6  &  78.9  &  79.3  &  75.6   \\
	Poly     &  75.4  &  73.0  &  73.5  &  72.8  &  75.9  &  75.2  &  76.4  &  74.6   \\
	ANN      &  75.9  &  72.7  &  73.6  &  74.6  &  77.1  &  76.6  &  77.8  &  75.5   \\
	\hline
	GCN      &  85.1  &  75.5  &  76.1  &  76.1  &  85.0  &  84.6  &  85.5  &  81.1   \\
	GSage    &  84.9  &  75.9  &  75.3  &  74.6  &  85.6  &  85.2  &  84.8  &  80.9   \\
    GAT      &  85.5  &  76.5  &  75.4  &  78.0  &  87.0  &  85.7  &  86.0  &  82.0   \\
	\hline
	Net$^\text{\textbf{2f}}$  \
	         &  86.9  &  80.1  &  80.9  &  79.3  &  88.7  &  88.2  &  87.3  &  84.5   \\
	Net$^\text{\textbf{2a}}$  \
	         &  93.5  &  91.7  &  90.7  &  90.3  &  93.0  &  93.0  &  92.9  &  92.2   \\
	\hline
  \end{tabular}
  }
  \vspace{-3mm}
\end{table}

\subsection{Identifying Long Nets}

In practice, we are also interested in how different methods can identify those longer nets. Table \ref{tbl:10LongestNet} shows the accuracy in identifying the top 10\% longest nets. For each netlist, the 10\% longest nets are labeled as true, and the accuracy is measured in ROC curve's area under curve (AUC). Then, the accuracy for each design is averaged over its 10 netlists. Since the net with many cells tends to be long, we add a baseline which directly estimates net length with the number of cells. On average, ANN, Poly and ISPL outperform MC and the number of cells with AUC$\sim=$0.76. Models capturing two or more-hop neighbors outperform models viewing one-hop neighbors only. Then graph methods (AUC$\sim=$0.82) are better than ANN by learning with a larger receptive field reaching five-hop neighbors. By combining shallow and deep embeddings, Net$^\text{\textbf{2f}}$ outperforms normal GNN with AUC$\sim=$0.85. Net$^\text{\textbf{2a}}$  significantly outperforms all graph methods by learning the global information on edge features with its edge convolution layer.

\begin{figure}[!t]
  \centering
    \includegraphics[width=0.9\columnwidth]{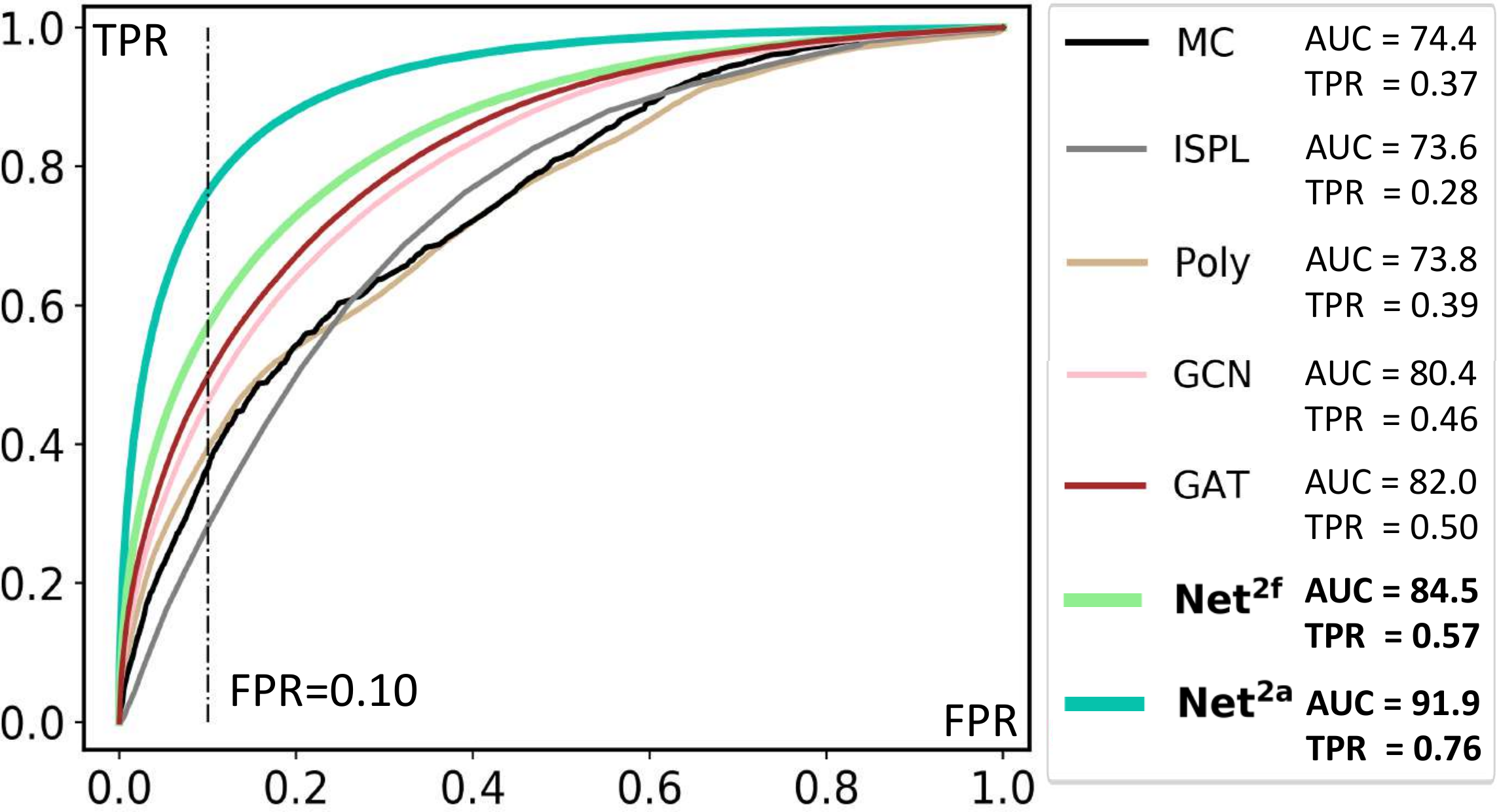}
      \vspace{-3mm}
  \caption{ROC curves in identifying 10\% longest nets.}
  \label{roc_curve}
\vspace{1mm}
\end{figure}

Figure \ref{roc_curve} shows a visualization of the ROC curves for identifying the top 10\% longest nets. A few less important methods are omitted for clarity. To visualize results from 70 netlists on one ROC curve, after prediction finishes, we put all nets from all 70 netlists together. Then the 10\% longest nets are selected and the ROC curve is measured on all these nets. The vertical dashed line means when the false positive rate (FPR) equals 0.1. The AUC measured on all nets for 10\% longest nets is fairly consistent with the average value shown in Table \ref{tbl:10LongestNet}. On the right of Figure \ref{roc_curve} also shows the true positive rate (TPR) for each method when FPR = 0.1. Both Net$^\text{\textbf{2f}}$ and Net$^\text{\textbf{2a}}$ significantly outperform previous works under all FPR.

\subsection{Results on Path Length Estimation}

In many EDA tools, the pre-placement timing report does not include wire delay, which hinders identifying critical paths accurately at an early stage. An  application of net length estimation is to predict the length of any given path, which correlates with post-placement wire delay on the path. The path length is defined as the summation of the net lengths over all nets on this path.

\begin{table}[!t]
% roc m:
  \vspace{-3mm}
  \centering
  \caption{Identify 10\% Longest Paths in ROC AUC (\%)}
  \vspace{-3mm}
  \label{tbl:10LongestPath}
  \resizebox{0.97\linewidth}{!}{
  \begin{tabular}{l | c c c c c c c | c}
 	\hline
 	\multirow{2}{*}{Methods}  &  \multirow{2}{*}{B14} &  \multirow{2}{*}{B15} &  \
 	 \multirow{2}{*}{B17} &  \multirow{2}{*}{B18} &   \multirow{2}{*}{B20} &  \
 	  \multirow{2}{*}{B21} &   \multirow{2}{*}{B22} &  \multirow{2}{*}{Ave} \\
 	  &     &     &     &     &     &     &     &     \\
	\hline
	%NumNet   &  54.7  &  54.5  &  65.0  &  68.5  &  70.0  &  59.0  &  69.0  &  63.0   \\
	ISPL     &  58.9  &  57.5  &  56.5  &  74.0  &  72.5  &  63.0  &  75.5  &  65.4   \\
	Poly     &  65.5  &  80.0  &  78.0  &  68.0  &  82.0  &  85.0  &  84.0  &  77.5   \\
	ANN      &  68.0  &  76.0  &  80.0  &  69.0  &  78.5  &  82.0  &  75.5  &  75.6  \\
	\hline
	GCN      &  63.5  &  75.0  &  86.5  &  56.0  &  82.0  &  81.5  &  85.5  &  75.7   \\
	GSage    &  65.0  &  88.0  &  93.0  &  77.0  &  81.5  &  67.0  &  80.0  &  78.8   \\
	GAT      &  63.0  &  92.0  &  95.0  &  83.5  &  83.5  &  76.0  &  89.5  &  83.2   \\
	\hline
	Net$^\text{\textbf{2f}}$  \
	         &  79.0  &  88.5  &  97.5  &  84.0  &  75.5  &  83.0  &  92.0  &  85.6   \\
	Net$^\text{\textbf{2a}}$  \
	         &  86.5  &  95.0  &  96.0  &  90.5  &  90.5  &  93.5  &  95.5  &  92.5   \\
	\hline
  \end{tabular}
  }
\end{table}

\begin{table}[!t]
\vspace{-5mm}
  \centering
  \caption{Comparing Pair of Paths by Lengths (\%)}
  \vspace{-3mm}
  \label{tbl:rankPath}
  \resizebox{0.97\linewidth}{!}{
  \begin{tabular}{l | c c c c c c c | c}
 	\hline
 	\multirow{2}{*}{Methods}  &  \multirow{2}{*}{B14} &  \multirow{2}{*}{B15} &  \
 	 \multirow{2}{*}{B17} &  \multirow{2}{*}{B18} &   \multirow{2}{*}{B20} &  \
 	  \multirow{2}{*}{B21} &   \multirow{2}{*}{B22} &  \multirow{2}{*}{Ave} \\
 	  &     &     &     &     &     &     &     &     \\
	\hline
	%NumNet   &  66.9  &  51.0  &  57.6  &  71.8  &  65.5  &  55.0  &  72.0  &  62.8   \\
	%NumCell  &  84.2  &  63.2  &  70.3  &  74.6  &  84.8  &  77.8  &  81.4  &  76.6   \\
	ISPL     &  67.1  &  55.0  &  58.2  &  77.4  &  68.9  &  59.7  &  69.5  &  65.1   \\
	Poly     &  83.9  &  86.6  &  83.3  &  70.4  &  83.4  &  80.4  &  86.3  &  82.0   \\
	ANN      &  82.0  &  74.8  &  75.3  &  68.1  &  81.9  &  65.4  &  80.5  &  75.4   \\
	\hline
	GCN      &  74.5  &  85.9  &  83.0  &  62.4  &  83.4  &  81.0  &  86.2  &  79.5   \\
	GSage    &  84.2  &  92.5  &  83.9  &  75.3  &  89.1  &  62.8  &  88.1  &  82.3   \\
	GAT      &  82.4  &  93.5  &  85.1  &  80.6  &  89.7  &  87.5  &  88.2  &  86.7   \\
	\hline
	Net$^\text{\textbf{2f}}$  \
	        &  87.3  &  92.7  &  87.6  &  93.1  &  91.1  &  91.2  &  86.9  &  90.0   \\
	Net$^\text{\textbf{2a}}$  \
	        &  96.8  &  97.0  &  91.4  &  95.9  &  92.2  &  94.2  &  94.4  &  94.6   \\
	\hline
  \end{tabular}
  }
  \vspace{1mm}
\end{table}

To verify models' performance on path length, for each netlist, we collect the timing-critical paths according to the pre-placement timing report. The report at this stage only includes cell delay. The path's slack has to be negative. Among these paths, the longer ones would be even more timing-critical considering wire delay after placement. Thus we apply models to identify those longest critical paths. The estimated path length is a summation over the predicted net lengths among all nets on the path.

Table \ref{tbl:10LongestPath} shows the ROC curve accuracy in identifying 10\% longest path. %We add a straightforward baseline named `NumNet', which means estimating path length with the number of nets on the path. Intuitively it should correlate with path length to a certain extent, but it turns out the weakest baseline. 
MC is not supposed to be \emph{linearly} proportional to actual wire length; thus the summation over MC is not included as path length estimation. Because the number of critical paths is much less than the number of nets, the trend on accuracy is not very consistent on different designs. On average, we can still observe an obvious trend that Net$^\text{\textbf{2a}}$ > Net$^\text{\textbf{2f}}$ > GAT > other methods.

In addition to locating the longest paths, we are interested in how models recognize the longer path when comparing any pair of two paths. To avoid meaningless comparisons between paths with very similar lengths, for each path, we compare it with all paths 30\% longer or shorter than it in its netlist. Table \ref{tbl:rankPath} shows the percentage of correct comparisons. Poly is the best method among previous works. The trend in accuracy is similar, Net$^\text{\textbf{2a}}$ > Net$^\text{\textbf{2f}}$ > GAT > other methods.

\subsection{Runtime Comparison}

\begin{table}[!b]
\vspace{1mm}
  \centering
  \caption{Runtime Comparison (In seconds)}
  \vspace{-3mm}
  \label{tbl:runtime}
  \resizebox{0.97\linewidth}{!}{
  \begin{tabular}{l | c | c  l  l  | c  c }
 	\hline
 	\multirow{2}{*}{Design}  &  \multirow{2}{*}{Place} \ 
 	& \multirow{2}{*}{Partition} & Net$^\text{\textbf{2f}}$  & Net$^\text{\textbf{2a}}$ \ 
 	& Net$^\text{\textbf{2f}}$   & Net$^\text{\textbf{2a}}$ \\
 	        &     &    &  Infer  & Infer  &  Speedup &  Speedup \\
	\hline 
 	Ave      &  97.8   &   7.0  &   0.05  &  0.07 &  1.7K $\times$  &  14.3$\times$   \\
	\hline
  \end{tabular}
  }
        \vspace{-3mm}
\end{table}

Table \ref{tbl:runtime} shows the runtime of placement and Net$^\text{\textbf{2}}$. The runtime is averaged over all netlists. For a fair comparison, the runtime of placement includes the placement algorithm only, without any extra time for file I/O, floorplanning, or placement optimization. Net$^\text{\textbf{2a}}$ takes slightly longer inference time than Net$^\text{\textbf{2f}}$ for its extra edge convolution layer. The overall runtime of Net$^\text{\textbf{2a}}$ includes both partition and inference. Partitioning takes the majority of the runtime. Net$^\text{\textbf{2a}}$ is more than 10$\times$ faster than placement.  The runtime of Net$^\text{\textbf{2a}}$ can be potentially improved by using coarser but faster partition $P$ and $M$, especially on larger designs. Without partition, Net$^\text{\textbf{2f}}$ is more than 1000$\times$ faster than placement. In addition, it takes around 30 minutes to train the Net$^\text{\textbf{2a}}$ model.

\section{Discussion}
\label{discussion}
%\subsection{Decompose Net$^\text{\textbf{2}}$}

In previous sections, we demonstrate the superior accuracy of Net$^\text{\textbf{2}}$ over other methods. Now we decompose Net$^\text{\textbf{2a}}$ to figure out which strategy contributes the most to its high accuracy. Table \ref{tbl:dis_10LongestNet} and \ref{tbl:dis_10LongestPath} measure the accuracy in identifying 10\% longest nets and 10\% longest paths. The `Edge ANN' means using edge features without GNN. Edge features are aggregated on their target node and processed with ANN together with node features. The `Simple Net' means using a simplified GNN structure. There is only one GAT layer for node convolution. The edge convolution only includes edge features and one layer of weights: $e_{k\_simple} = \sum_{b \in \mathcal{N}(k)} W_1 E_{b\rightarrow k}$. When not considering edge features, the receptive fields of `Edge ANN' and `Simple Net' are reduced to two hops and three hops, respectively. To analyze the contribution of different edge features, models `F0 Net', `F1 Net' and `F2+f\! 3 Net' use only $F0$, $F1$, $F2$\! $\&$\! $f3$ in edge features, respectively. The `Less $P$ Net' uses only the 4 coarser-granularity but faster cell partitions over the 7 partitions $P$ used in Net$^\text{\textbf{2a}}$. Thus its number of clusters equals the number of cells divided by 500, 1000, 2000, and 3000. Its smallest granularity is $5\times$ coarser than $P$. It is $\sim$1.5$\times$ faster than Net$^\text{\textbf{2a}}$ and more than 20$\times$ faster than placement.

All these variations from Net$^\text{\textbf{2a}}$ outperform GAT. The `Edge ANN' without GNN architecture is the worst variation, with around 4\% performance degradation. ANN's receptive filed and flexibility are not as good as Net$^\text{\textbf{2a}}$. But it still demonstrates the effectiveness of edge features.
The `Simple Net' is 1.5-2\% worse compared with Net$^\text{\textbf{2a}}$. Among the edge features, $F1$ contributes the most to accuracy. The model using less cell partitions $P$ is only around 0.6\% less accurate than Net$^\text{\textbf{2a}}$. It means using coarser-granularity partitions will not largely affect the performance of Net$^\text{\textbf{2a}}$.

\begin{table}[!tb]
%\vspace{-1mm}
  \centering
  \caption{Identify 10\% Longest Nets in ROC AUC (\%)}
  \vspace{-3mm}
  \label{tbl:dis_10LongestNet}
  \resizebox{\linewidth}{!}{
  \begin{tabular}{l | c c c c c c c | c}
 	\hline
 	\multirow{2}{*}{Methods}  &  \multirow{2}{*}{B14} &  \multirow{2}{*}{B15} &  \
 	 \multirow{2}{*}{B17} &  \multirow{2}{*}{B18} &   \multirow{2}{*}{B20} &  \
 	  \multirow{2}{*}{B21} &   \multirow{2}{*}{B22} &  \multirow{2}{*}{Ave} \\
 	  &     &     &     &     &     &     &     &     \\
	\hline
	Edge ANN   &  90.6  &  83.9  &  85.5  &  87.7  &  89.9  &  89.6  &  90.5  &  88.2   \\
	Simple Net &  92.3  &  89.6  &  88.9  &  89.7  &  91.6  &  91.7  &  91.7  &  90.8   \\
	\hline
	F0  Net    &  92.4  &  90.8  &  89.7  &  89.0  &  92.0  &  91.7  &  91.6  &  91.0   \\
	F1  Net    &  93.0  &  91.4  &  90.3  &  89.8  &  92.6  &  92.3  &  92.5  &  91.7   \\
	F2+f\! 3 Net    &  91.2  &  88.3  &  85.9  &  85.9  &  91.0  &  90.5  &  90.4  &  89.0   \\
	\hline
	Less $P$ Net &  92.9  &  91.2  &  89.7  &  89.6  &  92.7  &  92.5  &  92.3  &  91.6   \\
	Net$^\text{\textbf{2a}}$ \
	           &  93.5  &  91.7  &  90.7  &  90.3  &  93.0  &  93.0  &  92.9  &  92.2   \\
	\hline
  \end{tabular}
  }
      \vspace{1mm}
\end{table}

\begin{table}[!tb]
  \centering
  \caption{Identify 10\% Longest Paths in ROC AUC (\%)}
  \vspace{-3mm}
  \label{tbl:dis_10LongestPath}
  \resizebox{\linewidth}{!}{
  \begin{tabular}{l | c c c c c c c | c}
 	\hline
 	\multirow{2}{*}{Methods}  &  \multirow{2}{*}{B14} &  \multirow{2}{*}{B15} &  \
 	 \multirow{2}{*}{B17} &  \multirow{2}{*}{B18} &   \multirow{2}{*}{B20} &  \
 	  \multirow{2}{*}{B21} &   \multirow{2}{*}{B22} &  \multirow{2}{*}{Ave} \\
 	  &     &     &     &     &     &     &     &     \\
	\hline
	Edge ANN   &  77.0  &  91.0  &  94.5  &  88.0  &  85.0  &  89.5  &  94.0  &  88.4   \\
	Simple Net &  83.5  &  93.0  &  94.5  &  89.5  &  87.5  &  92.0  &  94.5  &  90.6   \\
	\hline
	F0 Net     &  77.0  &  89.5  &  95.0  &  90.0  &  90.0  &  87.5  &  93.5  &  88.9   \\
	F1  Net    &  81.5  &  93.0  &  95.5  &  90.5  &  89.5  &  87.5  &  95.0  &  90.4   \\
    F2+f\! 3  Net    &  82.0  &  92.5  &  97.0  &  85.5  &  87.5  &  93.5  &  93.5  &  90.2   \\
    \hline
    Less $P$ Net &  84.5  &  94.0  &  96.5  &  88.0  &  91.0  &  94.0  &  95.5  &  91.9   \\
    Net$^\text{\textbf{2a}}$ \
               &  86.5  &  95.0  &  96.0  &  90.5  &  90.5  &  93.5  &  95.5  &  92.5   \\
	\hline
  \end{tabular}
  }
    \vspace{2mm}
\end{table}

\section{Conclusion}

In this paper, we propose Net$^\text{\textbf{2}}$, a graph attention network method customized for individual net length estimation. It includes a fast version Net$^\text{\textbf{2f}}$ which is 1000 $\times$ faster than placement and an accuracy-centric version Net$^\text{\textbf{2a}}$ which efficiently extracts global information and outperform all previous methods. In future works, we will extend it to pre-placement timing analysis.

\section*{Acknowledgments}
This work is partially supported by Semiconductor Research Corporation Tasks 2810.021 and 2810.022 through UT Dallas’ Texas Analog Center of Excellence (TxACE).
%=======================================%
%=======================================%

%=======================================%
%               Bibliography            %
%=======================================%

\bibliographystyle{ACM-Reference-Format}
\nocite{kirby2019congestionnet}
\bibliography{PrePlace_zhiyao.bib}


\begin{thebibliography}{27}

%%% the Web address.

\ifx \showCODEN    \undefined \def \showCODEN     #1{\unskip}     \fi
\ifx \showDOI      \undefined \def \showDOI       #1{#1}\fi
\ifx \showISBNx    \undefined \def \showISBNx     #1{\unskip}     \fi
\ifx \showISBNxiii \undefined \def \showISBNxiii  #1{\unskip}     \fi
\ifx \showISSN     \undefined \def \showISSN      #1{\unskip}     \fi
\ifx \showLCCN     \undefined \def \showLCCN      #1{\unskip}     \fi
\ifx \shownote     \undefined \def \shownote      #1{#1}          \fi
\ifx \showarticletitle \undefined \def \showarticletitle #1{#1}   \fi
\ifx \showURL      \undefined \def \showURL       {\relax}        \fi
% invisible to TeX
\providecommand\bibfield[2]{#2}
\providecommand\bibinfo[2]{#2}
\providecommand\natexlab[1]{#1}
\providecommand\showeprint[2][]{arXiv:#2}

\bibitem[\protect\citeauthoryear{Bodapati and Najm}{Bodapati and Najm}{2001}]%
        {bodapati2001prelayout}
\bibfield{author}{\bibinfo{person}{Srinivas Bodapati} {and}
  \bibinfo{person}{Farid~N Najm}.} \bibinfo{year}{2001}\natexlab{}.
\newblock \showarticletitle{Prelayout estimation of individual wire lengths}.
\newblock \bibinfo{journal}{\emph{TVLSI}} (\bibinfo{year}{2001}).
\newblock


\bibitem[\protect\citeauthoryear{Cadence}{Cadence}{2020}]%
        {ispatial}
\bibfield{author}{\bibinfo{person}{Cadence}.} \bibinfo{year}{2020}\natexlab{}.
\newblock \bibinfo{title}{Cadence digital full flow optimized to deliver
  improved quality of results with up to 3X faster throughput}.
\newblock
\newblock
\urldef\tempurl%
\url{https://www.cadence.com/en_US/home/company/newsroom/press-releases/pr/2020/cadence-digital-full-flow-optimized-to-deliver-improved-quality-.html}
\showURL{%
\tempurl}


\bibitem[\protect\citeauthoryear{Corno, Reorda, and Squillero}{Corno
  et~al\mbox{.}}{2000}]%
        {corno2000rt}
\bibfield{author}{\bibinfo{person}{Fulvio Corno}, \bibinfo{person}{Matteo~Sonza
  Reorda}, {and} \bibinfo{person}{Giovanni Squillero}.}
  \bibinfo{year}{2000}\natexlab{}.
\newblock \showarticletitle{RT-level ITC'99 benchmarks and first ATPG results}.
\newblock \bibinfo{journal}{\emph{Design \& Test of computers}}
  (\bibinfo{year}{2000}).
\newblock


\bibitem[\protect\citeauthoryear{Fang et~al\mbox{.}}{Fang
  et~al\mbox{.}}{2018}]%
        {fang2018machine}
\bibfield{author}{\bibinfo{person}{Yen-Chun Fang} {et~al\mbox{.}}}
  \bibinfo{year}{2018}\natexlab{}.
\newblock \showarticletitle{Machine-learning-based dynamic IR drop prediction
  for ECO}. In \bibinfo{booktitle}{\emph{ICCAD}}.
\newblock


\bibitem[\protect\citeauthoryear{Fathi, Behjat, and Rakai}{Fathi
  et~al\mbox{.}}{2009}]%
        {fathi2009pre}
\bibfield{author}{\bibinfo{person}{Bahareh Fathi}, \bibinfo{person}{Laleh
  Behjat}, {and} \bibinfo{person}{Logan~M Rakai}.}
  \bibinfo{year}{2009}\natexlab{}.
\newblock \showarticletitle{A pre-placement net length estimation technique for
  mixed-size circuits}. In \bibinfo{booktitle}{\emph{SLIP}}.
\newblock


\bibitem[\protect\citeauthoryear{Fey and Lenssen}{Fey and Lenssen}{2019}]%
        {Fey/Lenssen/2019}
\bibfield{author}{\bibinfo{person}{Matthias Fey} {and} \bibinfo{person}{Jan~E.
  Lenssen}.} \bibinfo{year}{2019}\natexlab{}.
\newblock \showarticletitle{Fast graph representation learning with {PyTorch
  Geometric}}. In \bibinfo{booktitle}{\emph{ICLR-W}}.
\newblock


\bibitem[\protect\citeauthoryear{Hamilton, Ying, and Leskovec}{Hamilton
  et~al\mbox{.}}{2017}]%
        {hamilton2017inductive}
\bibfield{author}{\bibinfo{person}{Will Hamilton}, \bibinfo{person}{Zhitao
  Ying}, {and} \bibinfo{person}{Jure Leskovec}.}
  \bibinfo{year}{2017}\natexlab{}.
\newblock \showarticletitle{Inductive representation learning on large graphs}.
  In \bibinfo{booktitle}{\emph{NeurIPS}}.
\newblock


\bibitem[\protect\citeauthoryear{He, Zhang, Ren, and Sun}{He
  et~al\mbox{.}}{2016}]%
        {he2016deep}
\bibfield{author}{\bibinfo{person}{Kaiming He}, \bibinfo{person}{Xiangyu
  Zhang}, \bibinfo{person}{Shaoqing Ren}, {and} \bibinfo{person}{Jian Sun}.}
  \bibinfo{year}{2016}\natexlab{}.
\newblock \showarticletitle{Deep residual learning for image recognition}. In
  \bibinfo{booktitle}{\emph{CVPR}}.
\newblock


\bibitem[\protect\citeauthoryear{Hu and Marek-Sadowska}{Hu and
  Marek-Sadowska}{2003}]%
        {hu2003wire}
\bibfield{author}{\bibinfo{person}{Bo Hu} {and} \bibinfo{person}{Malgorzata
  Marek-Sadowska}.} \bibinfo{year}{2003}\natexlab{}.
\newblock \showarticletitle{Wire length prediction based clustering and its
  application in placement}. In \bibinfo{booktitle}{\emph{DAC}}.
\newblock


\bibitem[\protect\citeauthoryear{Hyun, Fan, and Shin}{Hyun
  et~al\mbox{.}}{2019}]%
        {hyun2019accurate}
\bibfield{author}{\bibinfo{person}{Daijoon Hyun}, \bibinfo{person}{Yuepeng
  Fan}, {and} \bibinfo{person}{Youngsoo Shin}.}
  \bibinfo{year}{2019}\natexlab{}.
\newblock \showarticletitle{Accurate wirelength prediction for placement-aware
  synthesis through machine learning}. In \bibinfo{booktitle}{\emph{DATE}}.
\newblock


\bibitem[\protect\citeauthoryear{Kahng and Reda}{Kahng and Reda}{2005}]%
        {kahng2005intrinsic}
\bibfield{author}{\bibinfo{person}{Andrew~B Kahng} {and}
  \bibinfo{person}{Sherief Reda}.} \bibinfo{year}{2005}\natexlab{}.
\newblock \showarticletitle{Intrinsic shortest path length: a new, accurate a
  priori wirelength estimator}. In \bibinfo{booktitle}{\emph{ICCAD}}.
\newblock


\bibitem[\protect\citeauthoryear{Karypis, Aggarwal, Kumar, and Shekhar}{Karypis
  et~al\mbox{.}}{1999}]%
        {karypis1999multilevel}
\bibfield{author}{\bibinfo{person}{George Karypis}, \bibinfo{person}{Rajat
  Aggarwal}, \bibinfo{person}{Vipin Kumar}, {and} \bibinfo{person}{Shashi
  Shekhar}.} \bibinfo{year}{1999}\natexlab{}.
\newblock \showarticletitle{Multilevel hypergraph partitioning: applications in
  VLSI domain}.
\newblock \bibinfo{journal}{\emph{TVLSI}} (\bibinfo{year}{1999}).
\newblock


\bibitem[\protect\citeauthoryear{Kipf and Welling}{Kipf and Welling}{2016}]%
        {kipf2016semi}
\bibfield{author}{\bibinfo{person}{Thomas~N Kipf} {and} \bibinfo{person}{Max
  Welling}.} \bibinfo{year}{2016}\natexlab{}.
\newblock \showarticletitle{Semi-supervised classification with graph
  convolutional networks}. In \bibinfo{booktitle}{\emph{ICLR}}.
\newblock


\bibitem[\protect\citeauthoryear{Kirby et~al\mbox{.}}{Kirby
  et~al\mbox{.}}{2019}]%
        {kirby2019congestionnet}
\bibfield{author}{\bibinfo{person}{Robert Kirby} {et~al\mbox{.}}}
  \bibinfo{year}{2019}\natexlab{}.
\newblock \showarticletitle{CongestionNet: Routing Congestion Prediction Using
  Deep Graph Neural Networks}. In \bibinfo{booktitle}{\emph{VLSI-SoC}}.
\newblock


\bibitem[\protect\citeauthoryear{Liu, Ma, and Zhang}{Liu et~al\mbox{.}}{2012}]%
        {liu2012neural}
\bibfield{author}{\bibinfo{person}{Qiang Liu}, \bibinfo{person}{Jianguo Ma},
  {and} \bibinfo{person}{Qijun Zhang}.} \bibinfo{year}{2012}\natexlab{}.
\newblock \showarticletitle{Neural network based pre-placement wirelength
  estimation}. In \bibinfo{booktitle}{\emph{FPT}}.
\newblock


\bibitem[\protect\citeauthoryear{Liu and Marek-Sadowska}{Liu and
  Marek-Sadowska}{2004}]%
        {liu2004pre}
\bibfield{author}{\bibinfo{person}{Qinghua Liu} {and}
  \bibinfo{person}{Malgorzata Marek-Sadowska}.}
  \bibinfo{year}{2004}\natexlab{}.
\newblock \showarticletitle{Pre-layout wire length and congestion estimation}.
  In \bibinfo{booktitle}{\emph{DAC}}.
\newblock


\bibitem[\protect\citeauthoryear{Ma et~al\mbox{.}}{Ma et~al\mbox{.}}{2019}]%
        {ma2019high}
\bibfield{author}{\bibinfo{person}{Yuzhe Ma} {et~al\mbox{.}}}
  \bibinfo{year}{2019}\natexlab{}.
\newblock \showarticletitle{High performance graph convolutional networks with
  applications in testability analysis}. In \bibinfo{booktitle}{\emph{DAC}}.
\newblock


\bibitem[\protect\citeauthoryear{Magen, Kolodny, Weiser, and Shamir}{Magen
  et~al\mbox{.}}{2004}]%
        {MagenSLIP04}
\bibfield{author}{\bibinfo{person}{Nir Magen}, \bibinfo{person}{Avinoam
  Kolodny}, \bibinfo{person}{Uri Weiser}, {and} \bibinfo{person}{Nachum
  Shamir}.} \bibinfo{year}{2004}\natexlab{}.
\newblock \showarticletitle{Interconnect-power dissipation in a
  microprocessor}. In \bibinfo{booktitle}{\emph{SLIP}}.
\newblock


\bibitem[\protect\citeauthoryear{Paszke et~al\mbox{.}}{Paszke
  et~al\mbox{.}}{2019}]%
        {paszke2019pytorch}
\bibfield{author}{\bibinfo{person}{Adam Paszke} {et~al\mbox{.}}}
  \bibinfo{year}{2019}\natexlab{}.
\newblock \showarticletitle{PyTorch: An imperative style, high-performance deep
  learning library}. In \bibinfo{booktitle}{\emph{NeurIPS}}.
\newblock


\bibitem[\protect\citeauthoryear{Pedram and Bhat}{Pedram and Bhat}{1991a}]%
        {PedramDAC91}
\bibfield{author}{\bibinfo{person}{Massoud Pedram} {and}
  \bibinfo{person}{Narasimha Bhat}.} \bibinfo{year}{1991}\natexlab{a}.
\newblock \showarticletitle{Layout driven technology mapping}. In
  \bibinfo{booktitle}{\emph{DAC}}.
\newblock


\bibitem[\protect\citeauthoryear{Pedram and Bhat}{Pedram and Bhat}{1991b}]%
        {PedramICCAD91}
\bibfield{author}{\bibinfo{person}{Massoud Pedram} {and}
  \bibinfo{person}{Narasimha~B Bhat}.} \bibinfo{year}{1991}\natexlab{b}.
\newblock \showarticletitle{Layout driven logic restructuring/decomposition}.
  In \bibinfo{booktitle}{\emph{ICCAD}}.
\newblock


\bibitem[\protect\citeauthoryear{Ronneberger, Fischer, and Brox}{Ronneberger
  et~al\mbox{.}}{2015}]%
        {ronneberger2015u}
\bibfield{author}{\bibinfo{person}{Olaf Ronneberger}, \bibinfo{person}{Philipp
  Fischer}, {and} \bibinfo{person}{Thomas Brox}.}
  \bibinfo{year}{2015}\natexlab{}.
\newblock \showarticletitle{U-net: Convolutional networks for biomedical image
  segmentation}. In \bibinfo{booktitle}{\emph{MICCAI}}.
\newblock


\bibitem[\protect\citeauthoryear{Synopsys}{Synopsys}{2020}]%
        {FusionCompiler}
\bibfield{author}{\bibinfo{person}{Synopsys}.} \bibinfo{year}{2020}\natexlab{}.
\newblock \bibinfo{title}{Fusion Compiler: the singular RTL-to-GDSII digital
  implementation solution}.
\newblock
\newblock
\urldef\tempurl%
\url{https://www.synopsys.com/implementation-and-signoff/physical-implementation/fusion-compiler.html}
\showURL{%
\tempurl}


\bibitem[\protect\citeauthoryear{Veli{\v{c}}kovi{\'c}
  et~al\mbox{.}}{Veli{\v{c}}kovi{\'c} et~al\mbox{.}}{2017}]%
        {velivckovic2017graph}
\bibfield{author}{\bibinfo{person}{Petar Veli{\v{c}}kovi{\'c}} {et~al\mbox{.}}}
  \bibinfo{year}{2017}\natexlab{}.
\newblock \showarticletitle{Graph attention networks}. In
  \bibinfo{booktitle}{\emph{ICLR}}.
\newblock


\bibitem[\protect\citeauthoryear{Xie et~al\mbox{.}}{Xie et~al\mbox{.}}{2018}]%
        {xie2018routenet}
\bibfield{author}{\bibinfo{person}{Zhiyao Xie} {et~al\mbox{.}}}
  \bibinfo{year}{2018}\natexlab{}.
\newblock \showarticletitle{RouteNet: Routability prediction for mixed-size
  designs using convolutional neural network}. In
  \bibinfo{booktitle}{\emph{ICCAD}}.
\newblock


\bibitem[\protect\citeauthoryear{Xie, Ren, Khailany, Sheng, Santosh, Hu, and
  Chen}{Xie et~al\mbox{.}}{2020}]%
        {xie2020powernet}
\bibfield{author}{\bibinfo{person}{Zhiyao Xie}, \bibinfo{person}{Haoxing Ren},
  \bibinfo{person}{Brucek Khailany}, \bibinfo{person}{Ye Sheng},
  \bibinfo{person}{Santosh Santosh}, \bibinfo{person}{Jiang Hu}, {and}
  \bibinfo{person}{Yiran Chen}.} \bibinfo{year}{2020}\natexlab{}.
\newblock \showarticletitle{PowerNet: Transferable Dynamic IR Drop Estimation
  via Maximum Convolutional Neural Network}. In
  \bibinfo{booktitle}{\emph{ASPDAC}}.
\newblock


\bibitem[\protect\citeauthoryear{Zhang, He, and Katabi}{Zhang
  et~al\mbox{.}}{2019}]%
        {zhang2019circuit}
\bibfield{author}{\bibinfo{person}{Guo Zhang}, \bibinfo{person}{Hao He}, {and}
  \bibinfo{person}{Dina Katabi}.} \bibinfo{year}{2019}\natexlab{}.
\newblock \showarticletitle{Circuit-GNN: Graph neural networks for distributed
  circuit design}. In \bibinfo{booktitle}{\emph{ICML}}.
\newblock


\end{thebibliography}
%=======================================%
%=======================================%
\end{document}